\documentclass[letterpaper]{article} 
\usepackage{aaai2026}  
\usepackage{times}  
\usepackage{helvet}  
\usepackage{courier}  
\usepackage[hyphens]{url}  
\usepackage{graphicx} 
\usepackage{natbib}  
\usepackage{caption} 
\usepackage{algorithm}
\usepackage{algorithmic}
\nocopyright
\usepackage{newfloat}
\usepackage{listings}

\usepackage{booktabs}

\DeclareCaptionStyle{ruled}{labelfont=normalfont,labelsep=colon,strut=off} 
\floatstyle{ruled}
\newfloat{listing}{tb}{lst}{}
\floatname{listing}{Listing}
\title{Automated Evolutionary Optimization for Resource-Efficient Neural Network Training}
\author{
    Ilia Revin\textsuperscript{\rm 1}, 
    Leon Strelkov\textsuperscript{\rm 1},
    Vadim A. Potemkin\textsuperscript{\rm 1}, \\
    Ivan Kireev\textsuperscript{\rm 2}, 
    Andrey Savchenko\textsuperscript{\rm 2},
}
\affiliations{
    \textsuperscript{\rm 1}ITMO University, Saint Petersburg, Russia\\
    \textsuperscript{\rm 2}SBER, Moscow, Russia\\

}

\usepackage{bibentry}

\begin{document}

\maketitle

\begin{abstract}
There are many critical challenges in optimizing neural network models, including distributed computing, compression techniques, and efficient training, regardless of their application to specific tasks. Solving such problems is crucial because the need for scalable and resource-efficient models is increasing. To address these challenges, we have developed a new automated machine learning (AutoML) framework, Parameter Efficient Training with Robust Automation (PETRA). It applies evolutionary optimization to model architecture and training strategy. PETRA includes pruning, quantization, and loss regularization. Experimental studies on real-world data with financial event sequences, as well as image and time-series  –  benchmarks, demonstrate PETRA's ability to improve neural model performance and scalability  –  namely, a significant decrease in model size (up to 75\%) and latency (up to 33\%), and an increase in throughput (by 13\%) without noticeable degradation in the target metric.
\end{abstract}

\begin{links}
    \link{Code}{https://anonymous.4open.science/r/PETRA_experiments}
\end{links}

\section{Introduction}


Efficient and scalable training of neural networks remains a fundamental challenge in modern machine learning, especially in scenarios constrained by hardware limitations or real-time requirements \cite{abdelmoniem2023refl, yu2021toward, bai2024beyond}. Model efficiency is typically measured by inference latency, throughput, and memory footprint \cite{latency, liu2024ffsplit}. However, optimizing these metrics simultaneously without sacrificing model quality is a complex and computationally demanding task.


One promising direction is parameter-efficient fine-tuning (PEFT), which adapts pre-trained models to downstream tasks by modifying only a small subset of their parameters \cite{runwal2024peft, han2024parameterefficientfinetuninglargemodels}. While PEFT methods can significantly reduce training costs, their practical use often requires manual configuration of modules and hyperparameters, limiting scalability and automation. Frameworks like AutoPEFT \cite{zhou2024autopeft} attempt to automate this process via Bayesian optimization, but are narrowly scoped to large language models (LLMs) and cannot be easily generalized to other domains.


In this paper, we introduce PETRA (Parameter-Efficient Training with Robust Automation), a domain-general AutoML framework designed to automatically construct efficient training pipelines using evolutionary optimization. PETRA explores the space of model compression and fine-tuning techniques – including pruning, quantization, and loss regularization – using a multi-objective strategy that balances model quality with computational efficiency \cite{hao2024modeluncertaintyevolutionaryoptimization}. Unlike existing methods, PETRA is applicable across model families and domains. We validate its generality and effectiveness on a diverse set of benchmarks, including financial time series, image classification, and energy consumption prediction tasks.

\section{Related works}

\subsection{AutoML frameworks}

Automated machine learning (AutoML) aims to reduce user involvement in developing machine learning models by automating model selection, hyperparameter tuning, and preprocessing \cite{karmaker2021automl, alsharef2022review}. Popular frameworks include TPOT \cite{olson2016tpot}, H2O \cite{ledell2020h2o}, Fedot \cite{10254012}, LightAutoML \cite{vakhrushev2021lightautoml}, and AutoGluon \cite{qi2021autogluon}.

Several systems employ evolutionary optimization to navigate flexible search spaces. For example, Fedot \cite{nikitin2021automated} explores atomic model compositions, while Fedot.Industrial \cite{revin2023automated} focuses on time series transformations. Fedot.Industrial extends this by adaptively composing pipelines, showing competitive performance in time series classification, regression, and forecasting.

However, most AutoML tools target classical ML or treat deep models as black boxes, rarely addressing model compression or parameter-efficient strategies. To our knowledge, no AutoML framework integrates PEFT methods – e.g., pruning, quantization, structured regularization – across diverse domains. AutoPEFT \cite{zhou2024autopeft} addresses PEFT using Bayesian optimization, but is limited to large language models (LLMs).

Our work fills this gap by embedding PEFT modules into a domain-general AutoML framework guided by multi-objective evolutionary search. PETRA treats compression techniques as core pipeline components, enabling efficient training and deployment across architectures and tasks.

\subsection{Parameter-efficient training methods}

Parameter-efficient fine-tuning (PEFT) techniques aim to reduce the number of trainable parameters required to adapt pre-trained models across tasks. Recent advances demonstrate improved performance and efficiency in diverse domains. For instance, FreqFit applies LoRA and Adapter modules in the frequency domain to enhance pattern recognition \cite{ly2024enhancing}, while Point-PEFT targets 3D point cloud classification using minimal parameter updates \cite{tang2024point}. Gradient-based Parameter Selection (GPS) selectively fine-tunes parameters based on gradient importance scores, achieving competitive performance in various tasks \cite{zhang2024gradient}.

Classical approaches such as Adapters \cite{han2024parameterefficientfinetuninglargemodels}, LoRA \cite{lin2025tracking}, and Prefix-Tuning \cite{kim2024preserving} freeze most model weights while inserting lightweight task-specific modules. These techniques have been extended to multi-modal architectures and dynamic configurations \cite{mao2021unipelt}, but typically require manual tuning of modules and hyperparameters. Automated PEFT configuration is a promising direction to address this limitation. AutoPEFT \cite{zhou2024autopeft} introduced a multi-objective Bayesian optimization framework to discover optimal module combinations, though it is limited to large language models.

Integrating PEFT into general-purpose AutoML systems presents new challenges, such as selecting pruning strategies and balancing regularization with architecture-specific constraints. In this work, we incorporate key PEFT strategies – pruning, quantization, and structured regularization – into a unified evolutionary optimization framework that automates the design of efficient training pipelines with minimal user input.

\section{Proposed approach}

In this paper, we propose an approach that adapts classical compression and fine-tuning techniques to the setting of automated, parameter-efficient training. The core idea is to represent neural network training pipelines as individuals in an evolutionary optimization process. The search space consists of neural networks and a set of operations on them – namely, pruning, quantization, and low-rank decomposition.

Each individual corresponds to a parameter-efficient training pipeline applied to an initial model. The pipeline's components and configuration serve as its genetic features, while its evaluation metrics guide selection.

Formally, the optimization objective is defined as:

\begin{equation}
\label{f:best_model}
    {M}_{opt}^{P} = {P}_{opt}({M}_{init}),
\end{equation}
\begin{equation}
\label{f:best_pipeline}
    {P}_{opt} = argmax_{P}{F(Q({M}^{P}), C({M}^{P}), S({M}^{P}))},
\end{equation}
\begin{equation}
\label{f:pipeline}
    P = (G, {\{{H}_{node}\}}_{G}, \{{H}_{M}\}),
\end{equation}

Here, $M_{opt}$ is the model produced by applying pipeline $P$ to the base model $M_{init}$. The function $Q(\cdot)$ denotes a quality-based criterion (e.g., ROC-AUC), $C(\cdot)$ represents computational metrics (e.g., latency, throughput), and $S(\cdot)$ reflects structural complexity constraints, such as pipeline depth or training time. These are jointly optimized via a Pareto hypervolume-based objective function $F(\cdot)$. Metrics like latency and model size are negated to align with the maximization objective. Each pipeline $P$ is described by a graph structure $G$, hyperparameters for each training stage $H_{node}$, and overall model-level parameters $H_M$.

A distinguishing feature of PETRA compared to AutoPEFT \cite{zhou2024autopeft} lies in its strategy for generating offspring. Since architectural differences between pipelines make crossover operations ill-defined, PETRA relies entirely on mutation-based variation. These mutations fall into two categories:

\begin{itemize}
  \item \textbf{Local mutations}, which modify internal hyperparameters of a specific PEFT node (e.g., pruning ratio, rank selection).
  \item \textbf{Global mutations}, which adjust outer components of the pipeline, such as the choice of optimizer or loss function.
\end{itemize}

This mutation-driven design enables flexible, fine-grained exploration of the pipeline search space without requiring handcrafted templates. Further details on regularization techniques and mutation operations are presented in the following sections.


\subsection{Loss Regularization and Low-Rank Decomposition}

Regularization is essential in deep learning for improving generalization and reducing overfitting. It typically involves three strategies: modifying the loss function, adjusting the network architecture (e.g., dropout, batch normalization), and applying data-driven techniques like augmentation. However, these approaches often slow rather than prevent overfitting \cite{zhang2021understanding}, and even standard optimizers like SGD can outperform accelerated ones in terms of generalization \cite{jacot2018neural}.

This effect is partially explained by the Neural Tangent Kernel (NTK) regime, where deep networks tend to converge to low-rank solutions. Accordingly, low-rank decomposition has proven effective for reducing model size and computational cost by approximating weight matrices with lower-rank representations \cite{yu2017compressing, hu2021lora}.

In our framework, we integrate regularization with low-rank approximation using singular value decomposition (SVD), applied to linear, convolutional, and embedding layers. We evaluate rank selection through criteria such as energy, explained variance, and singular value proportion. To guide convergence toward low-rank structure, we introduce two regularization terms:

\begin{equation}
\label{f:ortho_loss} 
    L_{O}(U, V) = \frac{1}{r^2}(|| U^TU - I || ^2_F + || V^TV - I ||^2_F),
\end{equation}
and Hoer loss:
\begin{equation}
\label{f:hoer_lss}
    L_H(S) = \frac{|| S ||_1}{|| S ||_2} = \frac{\sum_i |s_i|}{\sqrt{\sum_i{s_i^2}}}
\end{equation}
averaged by all layers which are decomposable with singular value decomposition as shown in the following equation:
\begin{equation}
\label{f:total_loss}
    L = L_{train} +\frac{\lambda_O}{|D|}\sum_{d \in D} L_O(U_d, V_d) + \frac{\lambda_{H}}{|D|}\sum_{d \in D} L_H(S_d).
\end{equation}

where $D$ is the set of SVD-decomposed layers, and $\lambda_O, \lambda_H$ are regularization weights.

To further enhance regularization, PETRA incorporates:

\begin{itemize}
    \item Lai Loss \cite{lai2024loss}, which stabilizes gradient flow by constraining its magnitude.
    \item A sparsity-inducing term \cite{bonetta2022regpruning} that promotes zeroing of irrelevant weights for effective pruning.
    \item Norm Loss \cite{georgiou2021normloss}, which blends norm-based objectives to encourage sparsity while retaining key weights.
\end{itemize}

Combined with low-rank decomposition, these techniques form a unified strategy that improves model compression, stability, and generalization across diverse architectures.

\subsection{Pruning and Quantization}

Pruning and quantization are widely used model compression techniques that significantly reduce storage, improve inference speed, and enable deployment on resource-constrained hardware \cite{li2023model, liang2021pruning}. Pruning eliminates redundant parameters – weights, neurons, or entire layers – while maintaining acceptable model performance. PETRA implements multiple importance-based pruning criteria, including magnitude, Taylor expansion \cite{molchanov2019importance}, Hessian sensitivity \cite{NIPS1989_6c9882bb}, batch norm scaling \cite{liu2017learning}, and LAMP \cite{lee2010layer}, applied across all layers.

Quantization further compresses models by reducing numerical precision. PETRA supports three quantization modes: post-training static (PTQ), post-training dynamic (PDQ), and quantization-aware training (QAT), with automatic selection during pipeline evolution. Weight types (e.g., INT8, FP16) are chosen depending on the target hardware. Recent studies show that converting to 8-bit weights reduces latency and improves throughput by 2–3x with minimal accuracy drop (1–2\%) \cite{10.3389/frobt.2025.1518965}. Since not all devices support INT8, PETRA defaults to FP16 quantization on GPUs.

When used together, pruning and quantization can produce highly compact models with negligible quality loss. For instance, combined strategies can reduce model size by 35x and increase throughput by 3–4x without accuracy degradation \cite{han2016deepcompressioncompressingdeep}. In PETRA, these methods are treated as modular components within the search space, allowing the evolutionary algorithm to identify the optimal compression configuration automatically.

\subsection{PETRA framework}

The PETRA framework addresses multi-objective optimization for parameter-efficient model training by evolving training pipelines through a population-based evolutionary algorithm. Each pipeline combines efficiency-enhancing modules (e.g., pruning, quantization, low-rank decomposition) and is evaluated based on accuracy, computational cost, and structural complexity.

\begin{figure*}[htbp]
    \centering
    \includegraphics[width=\linewidth]{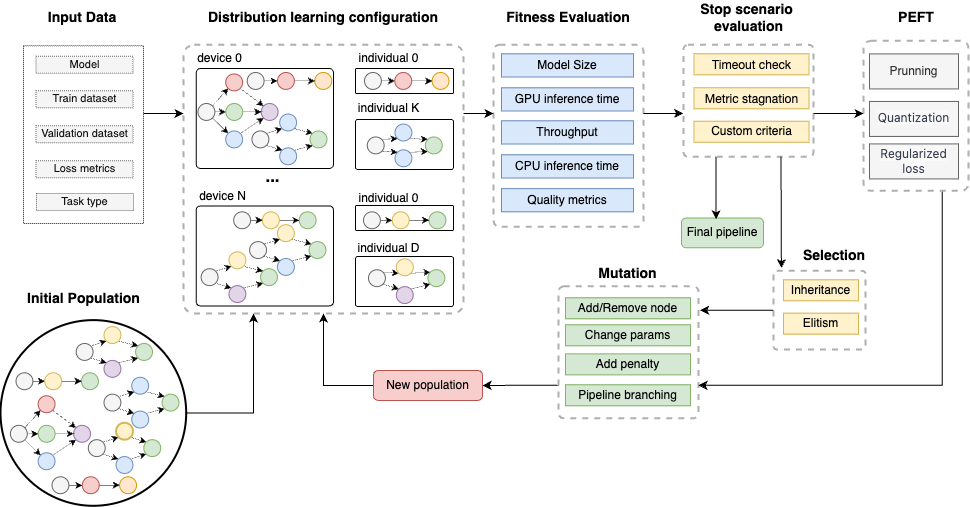}
    \caption{The proposed PETRA framework}
    \label{fig:petra}
\end{figure*}

As shown in Figure~\ref{fig:petra}, the process begins with an initial model built using the specified training and validation data, loss functions, and task type. This model is transformed into an initial population of $N$ pipelines, distributed across available computational resources.

Evolution proceeds via mutation-based operators that modify pipeline components such as training strategies, network architecture, loss terms, and optimizer settings. After each mutation, a new candidate is evaluated, and the Pareto front is updated based on objective values. To prevent population stagnation, PETRA adapts mutation probabilities based on the success rate of operator applications. Candidate selection combines Pareto-optimal pipelines with randomly sampled individuals to preserve diversity.

Two operational modes are supported:
\begin{itemize}
  \item \textbf{Pretrained model initialization}: the initial population is seeded with $N$ independently configured pipelines using a fixed base model.
  \item \textbf{Untrained model initialization}: pipelines are generated atomically and trained in parallel to maximize hardware utilization.
\end{itemize}

To reduce overhead, PETRA incorporates checkpoint reuse for comparing pipeline variants and applies an early-stopping mechanism (DepthAdaptation \cite{DBLP:journals/corr/abs-2103-01301}) to prune ineffective pipelines during training.

The evolutionary cycle continues until a stopping criterion is met – such as time, number of generations, or a target performance threshold – yielding a set of Pareto-optimal pipelines that balance model quality and resource efficiency across deployment settings.

\section{Experimental study}


To demonstrate the generalizability of the developed method, we conducted experiments on a diverse set of datasets and neural architectures spanning multiple domains. Specifically, we applied PETRA to models used for time-series regression, image classification, and financial prediction tasks.

For time-series related task, we employed deep convolutional architecture InceptionTime. For financial event sequence modeling, we used CoLES \cite{babaev2022coles}, a contrastive unsupervised learning model, in combination with LightGBM for downstream classification. In the case of image classification, standard benchmarks were used with ResNet model pretrained for 200 epochs before applying PETRA.

In the CoLES pipeline, the sequence encoder is first pre-trained using contrastive self-supervision, followed by 30 epochs of conventional training. PETRA is then applied to optimize the encoder. The learned embeddings are used to train a LightGBM classifier to evaluate representation quality.

\subsection{Datasets}

\textbf{Alpha Battle}~\cite{alphabattle}. The anonymized credit card transaction dataset is designed to analyze transactions and predict credit product default. The probability of default is estimated based on the history of consumer behavior in card transactions. The initial data include the currency type, transaction volume, and historical transaction time data. The training sample size is 4,326,918 transactions. The test sample size is 1,081,730 transactions.

\textbf{Age Group Prediction}~\cite{agepred}. The dataset of anonymized credit card transactions is designed to predict the probability that a customer belongs to a certain age segment based on the transaction data. The input data include the currency type, transaction volume, and historical transaction time data. The training sample size is 21,160,462 transactions. The test sample size is 5,290,115 transactions.

\textbf{CIFAR}~\cite{krizhevsky2009learning}. A collection of images commonly used to train machine learning and computer vision algorithms. In our case, we used the CIFAR-10 dataset.

\textbf{ImageNette}~\cite{ImageNet}. This is a publicly available large-scale database with annotated images, designed to be used in multiple computer vision tasks. It contains over 14 million images, but in our case, a limited version of Imagenette, which is a subset of 10 easily classified classes from ImageNet.

\textbf{Appliances Energy}~\cite{candanedo2017data}. A ZigBee wireless sensor network monitored house temperature and humidity conditions. Each wireless node transmitted the temperature and humidity conditions approximately every 3.3 minutes. Then, the wireless data was averaged over 10-minute periods. The energy data was logged every 10 minutes with m-bus energy meters. Weather data from the nearest airport weather station (Chievres Airport, Belgium) was downloaded from a public dataset from Reliable Prognosis (rp5.ru), and merged together with the experimental datasets using the date and time columns. Two random variables have been included in the dataset to test the  –  models and filter out non-predictive attributes (parameters).

\subsection{Equipment}


All experiments were conducted using an AMD EPYC 9124 16-Core processor and two NVIDIA RTX 6000 Ada Generation GPUs. This configuration allowed parallel evaluation of pipeline candidates during PETRA's optimization.

\subsection{Results and Discussion}





We evaluated PETRA using a multi-objective optimization setup that balances model quality, inference latency, throughput, and model size. The goal of this section is to answer the following high-level research questions:

\begin{itemize}
  \item \textbf{RQ1:} Can PETRA reduce model size and computational cost without significantly degrading predictive performance?
  \item \textbf{RQ2:} How does PETRA perform across different model architectures and domains (financial, image, time-series)?
  \item \textbf{RQ3:} What are the trade-offs between compression strategies in terms of latency, throughput, and model quality?
\end{itemize}

To evaluate these questions, we report results for each dataset/model pair, presenting pipelines from the Pareto front that offer distinct efficiency-quality trade-offs. Tables~\ref{tab:alpha_pipelines}–\ref{tab:timeseries-table} summarize the performance of selected pipelines relative to their original (uncompressed) models. The abbreviations used in the tables are as follows: Reg  –  Regularized Training, LR  –  Low-Rank Decomposition, Tr  –  Non-Regularized Training, Pr  –  Pruning, QAT  –  Quant-Aware Training, PDQ  –  Post-training Dynamic Quantization, PTQ  –  Post-Training Static Quantization, QD  –  Quantization Dynamic, QS  –  Quantization Static.

\textbf{Financial Data: Alpha Battle and Age Group:} Tables~\ref{tab:alpha_pipelines} and \ref{tab:age_pipelines} show the performance of PETRA-optimized pipelines on the Alpha Battle and Age Group datasets. In both cases, PETRA consistently reduced model size by 25–67\%, with minimal drop in ROC-AUC (1.6–4.6\% in the best pipelines). The best Alpha Battle pipeline decreased GPU latency by 12.5\% and increased GPU throughput by 10\%, with only a 1.8\% drop in ROC-AUC.

For the Age Group dataset, static quantization produced a 66.8\% reduction in model size and improved CPU throughput by 13.1\%, at the cost of a 2.6\% drop in ROC-AUC. These results show that PETRA can identify efficient configurations with a favorable balance between compression and performance.

\begin{table*}[htbp]
\centering
\caption{Pareto-optimal Individuals from Final Generation for Alpha Battle Dataset}
\label{tab:alpha_pipelines}
\resizebox{\textwidth}{!}{%
\begin{tabular}{lcccccc}
\toprule
\textbf{Pipeline} & \textbf{ROC-AUC} & \textbf{CPU Latency (ms)} & \textbf{GPU Latency (ms)} & \textbf{CPU Throughput (IPS)} & \textbf{GPU Throughput (IPS)} & \textbf{Model Size (MB)} \\
\midrule
Original & 0.770 & 0.239  & 0.0027 & 578  & 57144  & 18.646 \\
\cmidrule{1-7} 
Reg - LR - Tr - Pr & 0.741 / -3.7\% & 0.227 / -5.1\% & 0.0025 / -6.6\% & \textbf{600 / +3.9\%} & 61199 / +7.0\% & 15.311 / -17.9\% \\
Pr - QAT & 0.711 / -7.7\% & \underline{0.190 / -20.6\%} & $\infty$ & 586 / +1.4\% & $\infty$ & \underline{9.223 / -50.5\%} \\
Pr - Tr - Pr - PDQ & \underline{0.734 / -4.6\%} & 0.205 / -14.5\% & $\infty$ & \underline{602 / +4.1\%} & $\infty$ & 9.783 / -47.5\% \\
LR -Reg- Pr - LR & 0.726 / -5.7\% & \textbf{0.221 / -7.5\%} & \textbf{0.0023 / -13.7\%} & 592 / +2.4\% & \textbf{63822 / +11.7\%} & \textbf{13.802 / -26.0\%} \\
Reg - Pr - LR - Tr & \textbf{0.756 / -1.8\%} & 0.224 / -6.5\% & 0.0024 / -12.5\% & 590 / +2.0\% & 63109 / +10.0\% & 14.003 / -24.9\% \\
\bottomrule
\end{tabular}
}
\end{table*}

\begin{table*}[htbp]
\centering
\caption{Pareto-optimal Individuals from Final Generation for Age Group Dataset}
\label{tab:age_pipelines}
\resizebox{\textwidth}{!}{%
\begin{tabular}{lcccccc}
\toprule
\textbf{Pipeline} & \textbf{ROC-AUC} & \textbf{CPU Latency (ms)} & \textbf{GPU Latency (ms)} & \textbf{CPU Throughput (IPS)} & \textbf{GPU Throughput (IPS)} & \textbf{Model Size (MB)} \\
\midrule
Original & 0.621 & 0.299  & 0.0028 & 651 & 60134 & 1.710\\
\cmidrule{1-7} 
Pr - LR & \textbf{0.605 / -2.6\%} & 0.288 / -3.5\% & 0.0029 / -3.4\% & \textbf{681 / +4.6\%} & \textbf{67500 / +12.3\%} & 1.485 / -13.2\% \\
LR -Reg- Pr - LR & 0.600 / -3.4\% & 0.290 / -3.1\% & \textbf{0.0028 / -4.7\%} & 670 / +2.9\% & 66662 / +10.9\% & 1.387 / -18.9\% \\
Reg - LR & 0.593 / -4.7\% & \textbf{0.285 / -4.8\%} & 0.0029 / -1.3\% & 677 / +3.9\% & 67088 / +11.6\% & \textbf{1.385 / -19.0\%} \\
LR - PTQ & \underline{0.611 / -1.6\%} & \underline{0.201 / -32.7\%} & $\infty$ & \underline{736 / +13.1\%} & $\infty$ & 0.584 / -65.9\% \\
Pr - PDQ & 0.594 / -4.4\% & 0.238 / -20.4\% & $\infty$ & 712 / +9.3\% & $\infty$ & \underline{0.567 / -66.8\%} \\
\bottomrule
\end{tabular}
}
\end{table*}

\textbf{Image Classification: CIFAR-10 and ImageNette:} Tables~\ref{tab:cifar-table} and \ref{tab:imagenette-table} report results for image classification benchmarks using the ResNet model. For CIFAR-10, PETRA produced pipelines with up to 75\% reduction in model size. The most effective pipeline preserved F1 score ($<0.1\%$ drop) while increasing GPU throughput and slightly improving latency.

On ImageNette, PETRA achieved model compression of 76.6\%, and one configuration (LR–QAT–Pr–QS) even improved the F1 score by 4.5\% compared to the original. These results highlight PETRA's ability to maintain or improve predictive quality in computer vision tasks while reducing computational cost.

\begin{table*}[htbp]
\centering
\caption{Results by pipelines on CIFAR dataset and ResNet model}
\label{tab:cifar-table}
\resizebox{\textwidth}{!}{%
\begin{tabular}{lcccccc}
\toprule
\textbf{Pipeline} &
  \textbf{f1} &
  \textbf{CPU Latency (ms)} &
  \textbf{GPU Latency (ms)} &
  \textbf{CPU Throughput (IPS)} &
  \textbf{GPU Throughput (IPS)} &
  \textbf{Model Size (MB)} \\ \midrule
Original     & 0.759           & 0.004      & 1.901           & 1890      & 17           & 42.655           \\
LR - QD - Pr - QAT & 0.702 / -7.5\%  & $\infty$      & 2.271 / +19.4\% & $\infty$      & 15 / -11.7\% & 43.349 / +1.6\%    \\
LR - QS - Pr - QAT & 0.660 / -13.0\% & $\infty$      & 1.901 / +0\%    & $\infty$      & 17 / -0\%    & 10.854 / -74.6\% \\
\textbf{LR - QS - Pr - QD}  & \textbf{0.656 / -13.6}\% & $\infty$      & \textbf{1.573 / -17.3}\% & $\infty$      & \textbf{23 / +35.3}\%    & \textbf{10.861 / -74.5}\% \\
LR - QD - Pr - QAT & 0.758 / -0.1\%  & 0.005 / +25\% & $\infty$        & 1436 / -24\%  & $\infty$     & 44.137 / +3.5\%  \\
LR - QAT - Pr - QS & 0.747 / -1.6\%  & 0.005 / +25\% & $\infty$        & 1792 / -5.2\% & $\infty$     & 44.137 / +3.5\%  \\
LR - QS - Pr - QAT & 0.746 / -1.7\%  & 0.004 / 0\%   & $\infty$        & 1890 / -0\%   & $\infty$     & 44.198 / +3.6\%  \\ \bottomrule
\end{tabular}%
}
\end{table*}

\begin{figure*}[htbp]
    \centering
    \includegraphics[width=\textwidth]{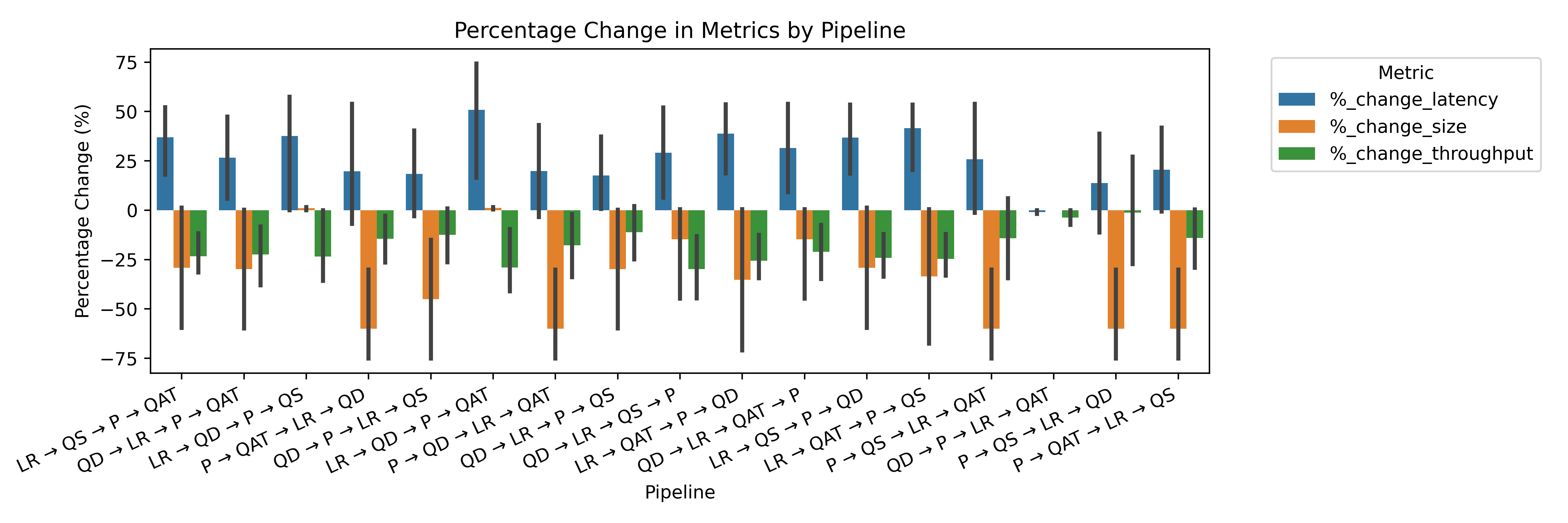}
    \caption{Percentage Change in Metrics by Pipeline for Model ResNet and CIFAR10 dataset}
    \label{fig:cifar_gpu}

\footnotesize
\vspace{0.2cm}
\end{figure*}

\begin{table*}[htbp]
\centering
\caption{Results by pipelines on Imagenette dataset and ResNet model}
\label{tab:imagenette-table}
\resizebox{\textwidth}{!}{%
\begin{tabular}{@{}lcccccc@{}}
\toprule
\textbf{Pipeline} &
  \textbf{f1} &
  \textbf{CPU Latency (ms)} &
  \textbf{GPU Latency (ms)} &
  \textbf{CPU Throughput (IPS)} &
  \textbf{GPU Throughput (IPS)} &
  \textbf{Model Size (MB)} \\ \midrule
Original     & 0.700           & 0.014     & 2.306           & 217     & 16           & 42.655           \\
LR - QAT - Pr - QS & 0.632 / -9\% & $\infty$     & 1.995 / -13.5\% & $\infty$     & 18 / +12.5\% & 9.960 / -76.6\%  \\
QD - Pr - QS - LR  & 0.500 / -28.6\%    & $\infty$     & 2.306 / -0\%    & $\infty$     & 16 / +0\%    & 10.453 / -75.5\% \\
LR - QAT - Pr - QD & 0.498 / -28.8\%  & $\infty$     & 1.948 / -15.5\% & $\infty$     & 18 / +12.5\% & 9.960 / -76.6\%  \\
\textbf{LR - QAT - Pr - QS} & \textbf{0.732 / +4.5}\% & \textbf{0.014 / +0}\% & $\infty$        & \textbf{208 / -4.1}\% & $\infty$     & \textbf{40.852 / -0}\%    \\
LR - QAT - Pr - QD & 0.709 / +1.3\% & 0.014 / +0\% & $\infty$        & 214 / -1.3\% & $\infty$     & 40.852 / -0\%    \\
LR - QD - Pr - QAT & 0.704 / +0.6\% & 0.013 / -7\% & $\infty$        & 226 / +4.1\% & $\infty$     & 40.852 / -0\%    \\ \bottomrule
\end{tabular}%
}
\end{table*}

\begin{figure*}[htbp]
    \centering
    \includegraphics[width=\textwidth]{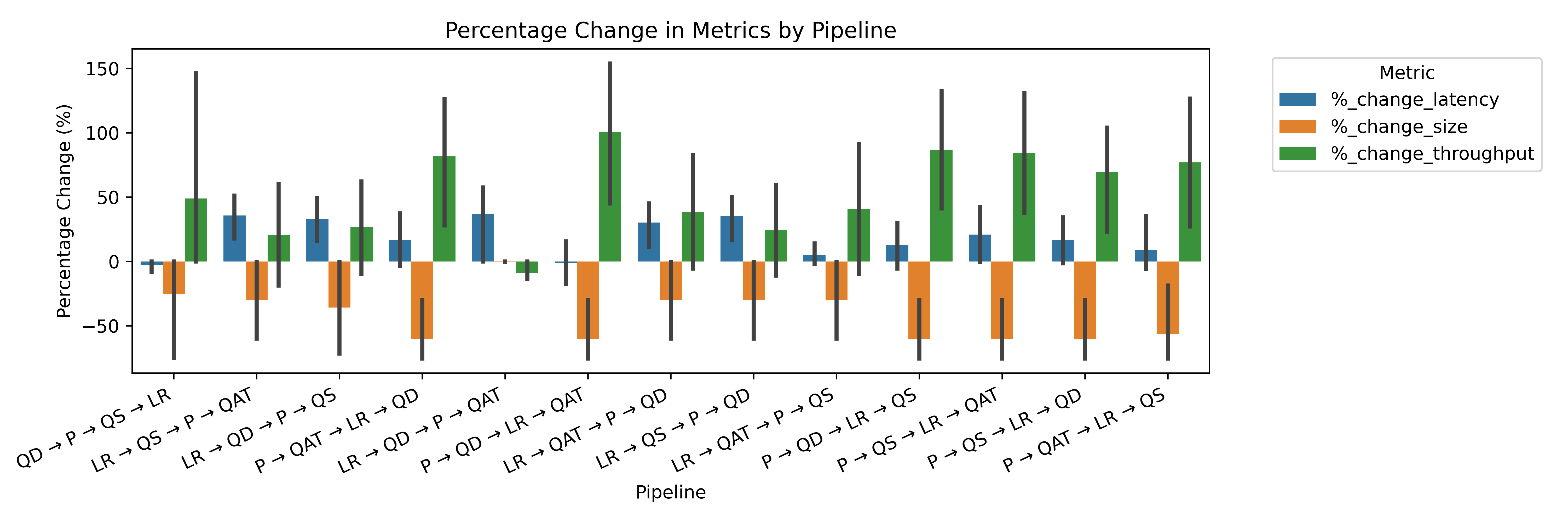}
    \caption{Percentage Change in Metrics by Pipeline for Model ResNet and ImageNette dataset}
    \label{fig:imagenet_gpu}

\footnotesize
\vspace{0.2cm}
\end{figure*}

\textbf{Time-Series Regression: Appliance Energy Dataset:} As shown in Table~\ref{tab:timeseries-table}, PETRA’s application to the InceptionTime model on the Appliance Energy dataset yielded mixed results. While model size was reduced by up to 85.2\%, RMSE increased significantly (up to +147\%). This suggests that PETRA's compression strategies, particularly quantization, may be less effective for time-series  –  where numerical precision is critical.

Moreover, certain pipelines led to reduced GPU throughput despite smaller models – indicating that aggressive compression can incur trade-offs when layer-level optimizations impact dataflow patterns unfavorably.

\begin{table*}[htbp]
\centering
\caption{Results by pipelines on ApplianceEnergy dataset and InceptionTime model}
\label{tab:timeseries-table}
\resizebox{\textwidth}{!}{%
\begin{tabular}{@{}lcccccc@{}}
\toprule
\textbf{Pipeline} &
  \textbf{rmse} &
  \textbf{CPU Latency (ms)} &
  \textbf{GPU Latency (ms)} &
  \textbf{CPU Throughput (IPS)} &
  \textbf{GPU Throughput (IPS)} &
  \textbf{Model Size (MB)} \\ \midrule
Original     & 3.441            & 0.009           & 4.345           & 175      & 2           & 2.954           \\
Pr - QS - LR - QD  & 3.811 / +10.8\%  & $\infty$        & 3.852 / -11.1\% & $\infty$      & 2.9 / +45\% & 0.923 / -68\%   \\
\textbf{Pr - QS - LR - QAT} & \textbf{3.817 / +10.9}\%  & $\infty$        & \textbf{3.538 / -18.5}\% & $\infty$      & \textbf{2.8 / -40}\% & \textbf{0.923 / -68}\%   \\
Pr - QAT - LR - QD & 3.841 / +11.6\%  & $\infty$        & 3.664 / -15.7\% & $\infty$      & 2.9 / -45\% & 0.923 / -68\%   \\
Pr - QAT - LR - QS & 7.611 / +123.1\% & 0.006 / -33.3\% & $\infty$        & 247 / -21.4\% & $\infty$    & 1.320 / -55.3\% \\
QAT - Pr - LR - QD & 8.002 / +134.5\% & 0.006 / -33.3\% & $\infty$        & 322 / -4.6\%  & $\infty$    & 0.437 / -85.2\% \\
Pr - QD - LR - QAT & 8.444 / +147.5\% & 0.007 / -22.2\% & $\infty$        & 257 / -28.7\% & $\infty$    & 1.694 / -42.7\% \\ \bottomrule
\end{tabular}%
}
\end{table*}

\begin{figure*}[htbp]
    \centering
    \includegraphics[width=\textwidth]{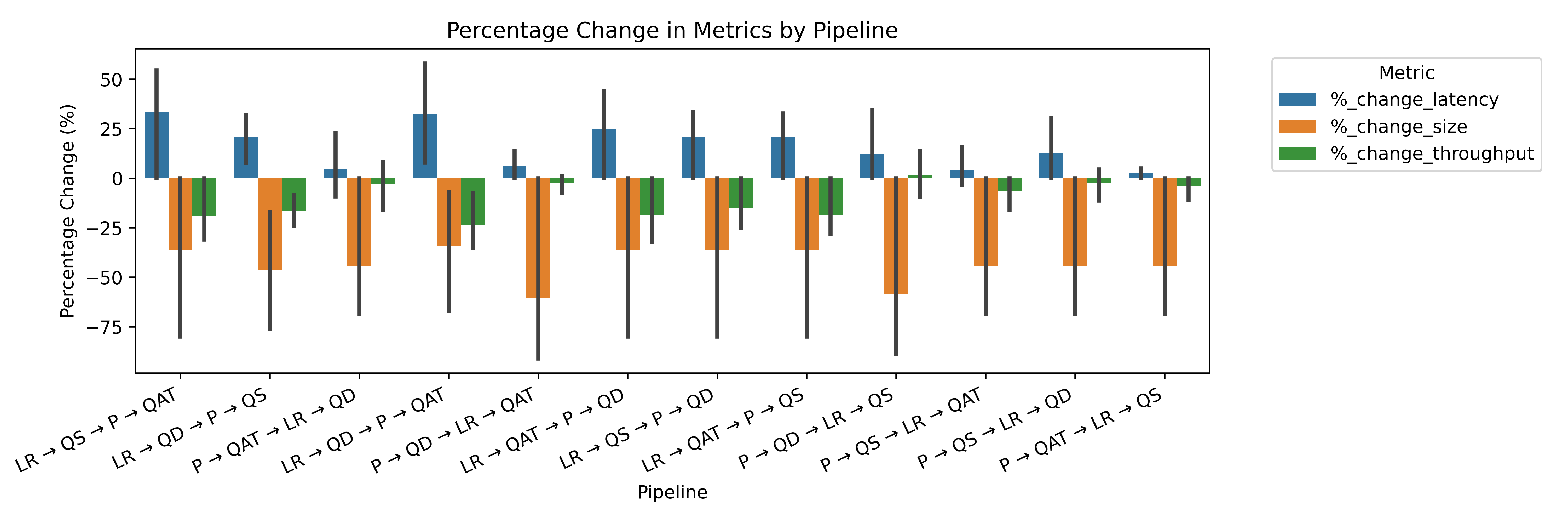}
    \caption{Percentage Change in Metrics by Pipeline for Model InceptionTime and ApplianceEnergy dataset}
    \label{fig:timeseries_gpu}

\end{figure*}

The following patterns were observed across datasets:

\begin{itemize}
  \item \textbf{Model Size:} Reductions of 25–85\% were achieved across all tasks. PETRA consistently discovered compact models with significant storage and memory benefits.
  \item \textbf{Latency and Throughput:} CPU/GPU latency dropped by up to 33\%, while throughput improved in most cases. However, gains were not universal and varied by architecture and pipeline design.
  \item \textbf{Accuracy/Quality:} In classification tasks, accuracy metrics (F1, ROC-AUC) typically decreased by less than 2–4\%. Larger drops occurred in  –  (RMSE), especially with aggressive compression.
\end{itemize}

PETRA demonstrates robust cross-domain performance, particularly for classification tasks where quantization and pruning introduce minimal degradation. Its evolutionary search effectively identifies Pareto-optimal trade-offs. However, in regression settings or tasks with precision-sensitive outputs, care must be taken to avoid over-compression.

Finally, while PETRA can produce highly compact models, some configurations may result in increased GPU latency due to quantization kernel overhead or inefficient scheduling – especially on deep architectures like InceptionTime.

\section{Conclusions}

We introduced PETRA, a domain-general AutoML framework for parameter-efficient neural network training. PETRA integrates model compression strategies – pruning, quantization, and loss-based regularization – into an evolutionary pipeline search process that automatically constructs training workflows optimized for both predictive quality and computational efficiency.

Experiments across financial classification, image recognition, and time-series regression tasks demonstrate that PETRA can achieve up to 85\% model size reduction, significant latency and throughput improvements, and minimal degradation in predictive metrics. In classification tasks, the accuracy drop was typically below 2--4\%, and in some cases, PETRA-optimized pipelines surpassed the baseline models.

These results support PETRA’s effectiveness as a general-purpose AutoML tool for producing compact, high-performing models adaptable to various deployment settings and hardware constraints.

\section*{Limitations}

Despite strong results across diverse tasks, PETRA has several limitations:

\begin{itemize}
  \item \textbf{Sensitivity in regression tasks:} On precision-sensitive problems such as time-series regression, aggressive compression can lead to notable performance degradation (e.g., RMSE increases), indicating the need for finer control over compression depth.
  
  \item \textbf{Latency unpredictability:} While PETRA generally reduces inference latency, certain configurations – especially those involving deep architectures like InceptionTime – may inadvertently increase latency due to inefficient layer-wise scheduling or memory access bottlenecks.

  \item \textbf{Search-time cost:} Although PETRA supports parallelization and includes early-stopping mechanisms such as DepthAdaptation, its evolutionary search process remains computationally demanding, especially for high-dimensional pipeline spaces.

\end{itemize}

Future work will focus on mitigating these limitations by incorporating hardware-aware search constraints, regression-specific adaptation strategies, and meta-learning techniques for guiding mutation and selection in high-complexity pipeline spaces.

\bibliography{aaai2026}


\end{document}